\documentclass{article}

    \usepackage[final]{iai_neurips_2024}

\usepackage[utf8]{inputenc} 
\usepackage[T1]{fontenc}    
\usepackage{hyperref}       
\usepackage{url}            
\usepackage{bm}
\usepackage{booktabs}       
\usepackage{amsfonts}       
\usepackage{nicefrac}       
\usepackage{microtype}      
\usepackage{xcolor}         
\usepackage{graphicx}
\usepackage{amsmath, amssymb}
\usepackage{algpseudocode}
\usepackage{algorithm}
\usepackage{algpseudocode}
\usepackage{pgfplots}
\usepackage{tikz}
\usepackage{pgfplots}
\usepackage{multirow}
\usepackage{booktabs}
\usepackage{pgfplots}
\pgfplotsset{compat=newest}
\usepgfplotslibrary{groupplots}  
\usepackage{tikz}
\usetikzlibrary{matrix, positioning}
\usepackage{hyperref}

\usepackage{amsmath}
\usepackage{float}

\DeclareMathOperator*{\argmin}{\arg\!\min}

\title{Riemann Sum Optimization for Accurate Integrated Gradients Computation}

\author{
  Swadesh Swain$^*$ \\
  Indian Institute of Technology, Roorkee\\
  \texttt{swadesh\_s@ece.iitr.ac.in} \\
   \And
  Shree Singhi$^*$ \\
  Indian Institute of Technology, Roorkee \\
  \texttt{shree\_s@mfs.iitr.ac.in}
}

\begin{document}

\maketitle

\def\thefootnote{*}\footnotetext{Equal Contribution}

\begingroup
\renewcommand{\thefootnote}{$\dagger$}
\footnotetext{Our code is available at \href{https://github.com/ShreeSinghi/RiemannOpt}{https://github.com/ShreeSinghi/RiemannOpt} }
\endgroup

\begin{abstract}
Integrated Gradients (IG) is a widely used algorithm for attributing the outputs of a deep neural network to its input features. Due to the absence of closed-form integrals for deep learning models, inaccurate Riemann Sum approximations are used to calculate IG. This often introduces undesirable errors in the form of high levels of noise, leading to false insights in the model's decision-making process. We introduce a framework, \textsc{\textbf{RiemannOpt}}$^\dag$, that minimizes these errors by optimizing the sample point selection for the Riemann Sum. Our algorithm is highly versatile and applicable to IG as well as its derivatives like Blur IG and Guided IG. \textsc{RiemannOpt} achieves up to $20\%$ improvement in Insertion Scores. Additionally, it enables its users to curtail computational costs by up to four folds, thereby making it highly functional for constrained environments.

\end{abstract}

\section{Introduction}
Deep Neural Network (DNN) classifiers for computer vision are increasingly being utilized in critical fields such as healthcare [\citenum{diabeetus}] and autonomous driving [\citenum{car}]. Hence, it has become increasingly important to understand the decision-making process for these models. This has led to a growing body of research focused on understanding how the predictions of these deep networks can be attributed to specific regions of the image. An attribution method attempts to explain which inputs the model considers to be most important for its outputs.  Several gradient-based [\citenum{smoothgrad, gradcam, Karen, xrai, lime, shapley}] and gradient-free [\citenum{lundb,rise, pmlr-v119-sundararajan20b, bbc, ruthfong, kis,springenberg2015strivingsimplicityconvolutionalnet,zintgraf2017visualizingdeepneuralnetwork}] attribution methods have been developed for deep learning models. Integrated Gradient methods [\citenum{blurig, gig, ig}] are a specific class of gradient-based attribution methods that compute a line integral of the gradients of the model over a path defined from a baseline image to the given input.

The complex functional space of deep learning models is often considered as a source of noise for many gradient-based attribution methods, resulting in undesirable high attribution to some background regions. For Integrated Gradients, \citet{gig} claim that the source of noise is large gradients in the model surface while \citet{smoothgrad} argue that the source of error is the rapid fluctuation of the gradients of deep learning models.

Deep learning models do not have closed-form integrals, so their integrals are approximated by the Riemann Sums [\citenum{ig}]. This approximation involves the sampling of a number of points along the path and approximating the integral using interpolation between these points. Using more points to approximate the Riemann sum naturally results in cleaner saliency maps. However, most applications of Integrated Gradients require a high number of steps for the Riemann Sum [\citenum{ig, 1000}], generally between $20$ to $1000$, rendering Integrated Gradients' usage practically unfeasible in real-time applications. On the other hand, using lesser number of samples severely impacts the quality of the saliency map. This results in a trade-off between speed and performance.

\citet{exactline} have attempted to tackle the above issue of inaccurate Integrated Gradients computation by exactly computing the underlying integral using \textsc{ExactLine}. However, their application is limited to neural networks that are composed of piece-wise linear operations. Most traditional models, like InceptionV3 [\citenum{inception}], ViT [\citenum{vit}] and ResNet [\citenum{resnet}], while being primarily composed of linear operations, also make use of non-linear operations like LayerNorm [\citenum{layernorm}], GroupNorm [\citenum{groupnorm}] and Attention [\citenum{attention}] thus prohibiting the use of \textsc{ExactLine}. Furthermore, the use of \textsc{ExactLine} might not be considered ideal in cases where there are computational constraints since it requires $\sim 14000$ gradient computations per image for large models.

To overcome the problem of redundancy by ineffective sampling schedules prevalent in Integrated Gradient methods, we introduce \textbf{\textsc{RiemannOpt}}, a framework to pre-determine optimal points for sampling to calculate Riemann Sums. The pre-determined points are specific only to the model. Hence, the computation to determine the points is only done once and does not need to be repeated for every image. Unlike \textsc{ExactLine}, our method does not impose any architectural constraints on the underlying model, with the additional benefit of requiring far fewer samples. 
We present qualitative and quantitative results for \textsc{RiemannOpt} on Integrated Gradients (IG) [\citenum{ig}], Blur Integrated Gradients (BlurIG) [\citenum{blurig}] and Guided IG (GIG) [\citenum{gig}]. Our method can be easily combined with existing IG-based methods and enable them to generate cleaner saliency maps.

\begin{figure}
    \centering
    \includegraphics[width=1\linewidth]{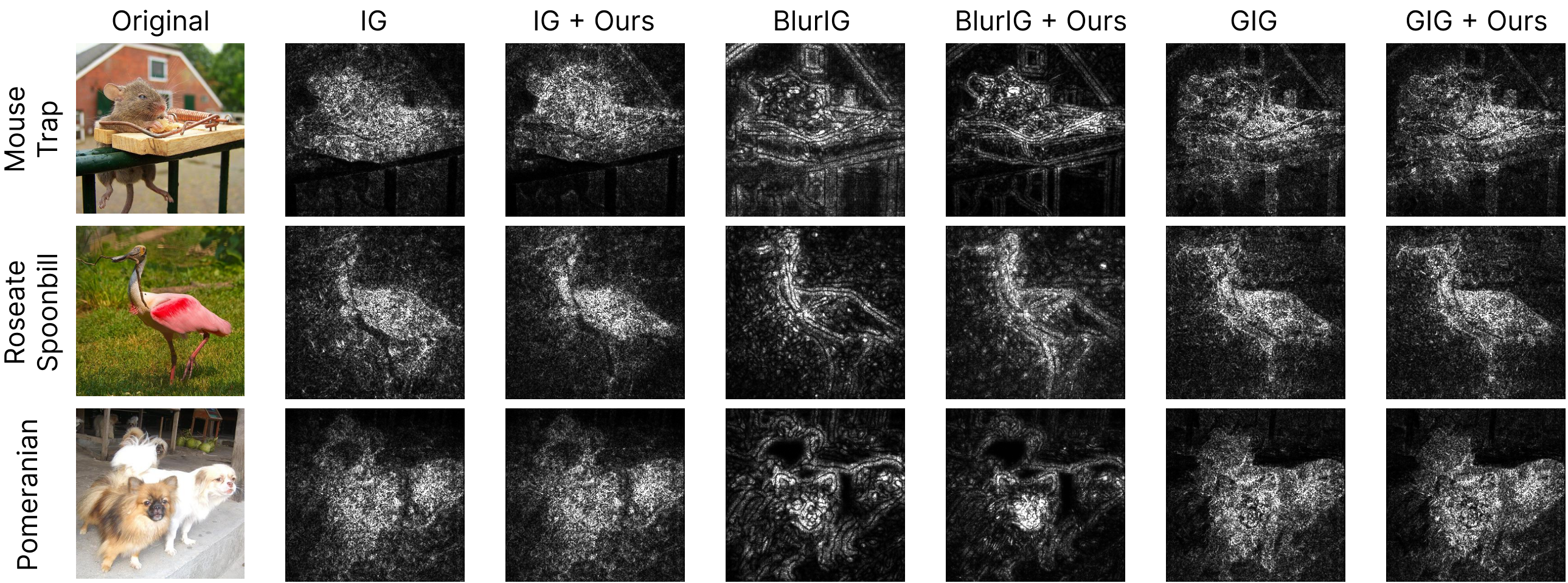}
    \caption{Visual comparison of Integrated Gradient methods with and without \textsc{RiemannOpt}. For IG, \textsc{RiemannOpt} suppresses the noise around the Spoonbill and also slightly concentrates stronger attribution scores on the mouse trap. Applying \textsc{RiemannOpt} to BlurIG significantly increases concentration on the subjects of images. GIG saliency maps remain perceptually similar.}
    \label{fig:enter-label}
\end{figure}

\section{Background}

In this section, we review the mathematical definition of IG [\citenum{ig}], BlurIG [\citenum{blurig}], and GIG [\citenum{gig}].

\subsection{Integrated Gradients}
\label{ig}

\citet{ig} utilized the idea of a path function. $\gamma : [0, 1] \rightarrow \mathbb{R}^n$ is a smooth function that denotes a path within $\mathbb{R}^n$ from $x'$ to $x$, satisfying $\gamma(0) = x'$ and $\gamma(1) = x$. Further, they defined path integrated gradients along the $i^{th}$ dimension for an input $x$, given a baseline $x'$, obtained by integrating the gradients along the path $\gamma(\alpha)$ for $\alpha \in [0, 1]$ as:

\begin{equation}
\label{ig_re}
        I_{i}(x)  =\int_{0}^{1}\frac{\partial f(\gamma(\alpha))}{\partial \gamma_{i}(\alpha)} \frac{\partial \gamma_{i}(\alpha)}{\partial \alpha} d \alpha, 
\end{equation}

Where $f$ denotes a DNN classifier. Integrated Gradients (IG) \citet{ig} originally defined the path method as a straight line path specified $\gamma^{IG}(\alpha) = x' + \alpha \times (x - x') \quad \text{for} \quad \alpha \in [0, 1]$. Later, BlurIG and GIG introduced non-linear paths that had their respective advantages over IG.

\subsection{Blur Integrated Gradients}
\label{blurig}

\citet{blurig} introduced Blur Integrated Gradients: For a given function \( f : \mathbb{R}^{m \times n} \to [0, 1] \) representing a classifier, let \( z(x, y) \) be the 2D input. Blur IG's path is defined by a Gaussian filter that progressively blurs the input. Formally:

\begin{equation}
\gamma^{BlurIG}(x, y, \alpha) = \sum_{m=-\infty}^{\infty} \sum_{n=-\infty}^{\infty} \frac{1}{\pi \alpha} e^{-\frac{x^2 + y^2}{\alpha}} z(x - m, y - n)
\end{equation}

The final BlurIG computation is as follows:

\begin{equation}
I^{BlurIG}(x, y) ::= \int_{\infty}^{0} \frac{\partial f_c(\gamma^{BlurIG}(x, y, \alpha))}{\partial \gamma^{BlurIG}(x, y, \alpha)} \frac{\partial \gamma^{BlurIG}(x, y, \alpha)}{\partial \alpha} d\alpha
\end{equation}

\subsection{Guided Integrated Gradients}
\label{gig}
Guided IG [\citenum{gig}] (GIG) follows an adaptive integration path $\gamma^{IG}(\alpha), \alpha \in [0,1]$ to avoid high gradient regions. An adaptive path is one that depends on the model being used:

\begin{equation}
    \gamma^{GIG} = \argmin_{\gamma \in \Gamma} \sum_{i=1}^N \int_{0}^{1} | \frac{\partial f(\gamma(\alpha))}{\partial \gamma_{i}(\alpha)} \frac{\partial \gamma_{i}(\alpha)}{\partial \alpha} | d \alpha,
\end{equation}

After finding the optimal path $\gamma^{GIG}$, GIG computes the attribution values similar to IG,

\begin{equation}
\label{gig_re}
        I_{i}^{GIG}(x)  =\int_{0}^{1}\frac{\partial f(\gamma^{GIG}(\alpha))}{\partial \gamma^{GIG}_{i}(\alpha)} \frac{\partial \gamma^{GIG}_{i}(\alpha)}{\partial \alpha} d \alpha.
\end{equation}

\section{Methodology}
In this section, we present simple derivation to determine an upper bound on the error introduced by approximating a one-dimensional integral using a Riemann Sum. We then extend the definition for multi-dimensional line integrals and define the algorithm \textsc{RiemannOpt} uses to schedule samples to minimize this upper bound.

\subsection{Error Minimization of Riemann Sums in 1D}
\label{derivation}
We now present the derivation to estimate the error introduced due to the left Riemann Sum approximation of a standard 1D integral $\int_{\alpha_0}^{\alpha_k} g(\alpha) \, d\alpha$ where  $\{\alpha_i\}_{i=0}^{k}$ is the set of points at which the integrand, $g(\alpha)$, is evaluated.
 
The standard way to calculate the left Riemann Sum is:

\begin{equation} \label{r_eq}
R = \sum_{i=0}^{n-1} g(\alpha_i) (\alpha_{i+1} - \alpha_i)
\end{equation}

The integral can be broken down as:
\begin{equation} \label{i_eq}
I = \sum_{i=0}^{n-1} \int_{\alpha_i}^{\alpha_{i+1}} g(\alpha) d\alpha
\end{equation}

By applying the Taylor Series approximation around $\alpha_{i}$ in \eqref{i_eq}:

\begin{equation}\label{i_eq2}
\begin{aligned}
I &\approx \sum_{i=0}^{n-1} \int_{\alpha_i}^{\alpha_{i+1}}  g(\alpha_i) + (\alpha-\alpha_i)g'(\alpha_i) \, dx \\
&\approx \sum_{i=0}^{n-1}  g(\alpha_i)(\alpha_{i+1}-\alpha_i) + g'(\alpha_i)\frac{(\alpha_{i+1}-\alpha_i)^2}{2}
\end{aligned}
\end{equation}

By \eqref{r_eq}, \eqref{i_eq2} and the Triangle Inequality:
\begin{equation}\label{final_result}
|R - I| \lesssim \frac{1}{2}\sum_{i=0}^{n-1} \left|g'(\alpha_i)(\alpha_{i+1}-\alpha_i)^2\right|
\end{equation}

\subsection{Algorithm}
\label{algorithm_reasoning}
IG computes $d$ integrals (attributions) per image, one for each pixel. We treat each integral independently and use the derivation above to estimate the average error over all integrals. The input to the function, $g$, would be multidimensional, resulting in a different Taylor Series expansion. However, the approximation would still be mathematically sound since the integral corresponding to the $i^{th}$ feature is only dependant on the gradient along that component, i.e. error corresponding to the $i^{th}$ dimension of gradient only contributes to the $i^{th}$ integral. We use this observation in conjunction with the finite distance approximation of the derivative to determine the optimal points for sampling a Riemann Sum for the dataset. The primary idea behind the algorithm is to approximate the average $|g'(\alpha)|$ for all input features on a small subset of images, $\sim 1\%$ of the validation dataset, then compute the optimal sampling points and use them for the entire dataset.

The following tensors are used in Algorithm~\ref{alg:optimal_alphas} where $d$ is the dimensionality of the input:
\begin{itemize}
    \item $I_{k \times d}$: Samples evaluated at $k$ equispaced points along the path.
    \item $C_{k-1 \times d}$: Finite difference estimate of the derivative of $I$ for all input features.
    \item $A_{k-1}$: Absolute derivative estimate of $I$, corresponding to $|g'(\alpha)|$
\end{itemize}

\algnewcommand{\Inputs}[1]{%
  \State \textbf{Inputs:}
  \Statex \hspace*{\algorithmicindent}\parbox[t]{.8\linewidth}{\raggedright #1}
}

\algnewcommand{\Outputs}[1]{%
  \State \textbf{Output:}
  \Statex \hspace*{\algorithmicindent}\parbox[t]{.8\linewidth}{\raggedright #1}
}

\algnewcommand{\Initialization}[1]{%
  \State \textbf{Initialization:}
  \Statex \hspace*{\algorithmicindent}\parbox[t]{.8\linewidth}{\raggedright #1}
}

\setlength{\textwidth}{\linewidth}

\begin{algorithm}
\caption{Estimation of Optimal Alphas \label{alg:optimal_alphas}}
\begin{algorithmic}

\Inputs{

A subset of $m$ examples from the validation dataset: $X_{i} \in \mathbb{R}^d, i \in \{{1, \ldots, m}\}$ \\
Number of sample points in a path: $k$ \\
Integrand of the IG method: $\frac{\partial f(\gamma(\alpha))}{\partial \gamma(\alpha)} \odot \frac{\partial \gamma(\alpha)}{\partial \alpha}$
} 

\Outputs{
Optimal sampling points: $\{\alpha^*_j\}_{j=1}^{k}$
}

\Initialization{
Set $\{\alpha_j\}_{j=1}^{k}$ as $k$ linearly spaced scalars between the integral bounds
\State $A \gets \text{Initialize with zeros}$
}
\newline

\For{\textbf{each } $i$ \textbf{ in } $\{1, \dots, m\}$} \Comment{Loop over training examples}
    \State $I_{j} \gets \frac{\partial f(\gamma(\alpha))}{\partial \gamma(\alpha)} \odot \frac{\partial \gamma(\alpha)}{\partial \alpha} \bigg\rvert_{\alpha = \alpha_j} \text{ for } j \text{ in } \{1, \dots, k\}$
    \State $C_{k-1 \times d} \gets \frac{I_{
    j+1} - I_{j}}{\alpha_{j+1} - \alpha_j} \text{ for } j \text{ in } \{1, \dots, k-1\}$ \Comment{Finite difference: $g'(\alpha) \approx \frac{g(\alpha+\Delta\alpha)-g(\alpha)}{\Delta\alpha}$}
    \State \text{Apply element-wise absolute to $C$}
    \State $A_{k-1} \mathrel{+}= \text{Average $C_{k-1 \times d}$ across all features}$ \Comment{Estimate of $|g'(\alpha)|$}
\EndFor

\State \text{Normalize $A$ by dividing by number of examples $m$}
\State $|g'(\alpha)| \gets \text{Linearly Interpolate}(A, \; \alpha)$ \Comment{$\alpha \in [0, 1], A \in \mathbb{R}^{k-1}$}

\State $\alpha^*_j \gets \text{The set $\{\alpha_j\}_{j=1}^{k}$ that minimizes the upper bound error defined by Equation \eqref{final_result}}$

\State \Return $\{\alpha^*_j\}_{j=1}^{k}$

\end{algorithmic}
\end{algorithm}

\section{Experimental Setup and Metrics}
In this section, we discuss the details of the implementation, dataset, model, and metrics used.

\subsection{Experimental Setup}
We use the original implementations with default parameters in the authors’ code for IG, GIG, and BlurIG and implement \textsc{RiemannOpt} as a pre-computation step that links with the original implementations. We present our results using InceptionV3 for 16, 32, 64 and 128 sample points on the correctly classified images of the ImageNet validation dataset, $\sim 40K$. To estimate $|g'(\alpha)|$, we apply Algorithm \ref{alg:optimal_alphas} to a set of $200$ randomly correctly classified images from the ImageNet validation dataset for 128 samples. Then, we use Powell's method [\citenum{powell}] to determine the optimal set of sampling points. This roughly has the same computational cost as computing the saliency map for the set of $200$ images. Using \textsc{RiemannOpt} is still cost-effective since we only use a small number of images to calculate sample points but are able to use these points for the entire dataset.

\subsection{Metrics}
Previous works use the Insertion Score and Normalized Insertion Score to compare different attribution methods [\citenum{blurig, xrai, gig, idgi, rise, pandeng}]. It is critical to note that the purpose of the Insertion Score is to measure the efficacy of a saliency map, i.e. it is not designed to measure how close the Riemann Sum is to the actual integral. However, it is reasonable to assume that the true saliency map would generally achieve better Insertion Scores than an inaccurate approximation since inaccurate estimates introduce noise. Hence, we report the Insertion Scores and, additionally, employ the Axiom of Completeness [\citenum{ig}] to define a new metric that measures the quality of the saliency maps without the need for this hypothesis.

According to the Axiom of Completeness, the sum of all feature attributions, determined by any Integrated Gradients method, must ideally add up to the difference between the output of $f$ at $x$ and $x'$. However, there is always an error due to inaccurate Riemann Sum estimates. Furthermore, \citet{ig} advise the developer to ensure that all feature attributions add up to $f(x) - f(x')$ (within 5\%) and suggest increasing the number of samples if the error is greater.

Since the ground truth is unavailable, it is non-trivial to determine the numerical accuracy of a computed saliency map. We use the relative error between the sum of feature attributions and $f(x) - f(x')$ to estimate the error. This metric is not infallible since the features' positive and negative errors partially offset each other during the summation. Using the Triangle Inequality, it can be easily shown that this metric is a lower bound on the true error. Nevertheless, it serves as a helpful proxy since near-perfect saliency maps will have near-zero error, and highly erroneous maps will, on average, have high error even after the errors partial offset.

\begin{figure}[H]
    \centering
    \includegraphics[width=1\linewidth]{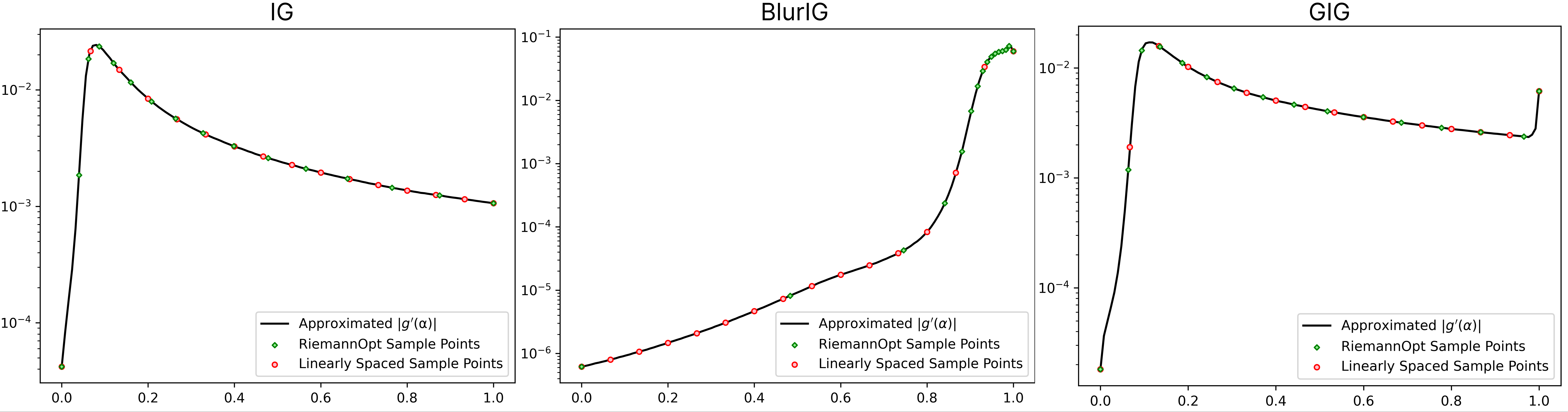}
    \caption{Estimated $|g'(\alpha)|$ and comparison of 16 linearly spaced samples and 16 optimal samples chosen by \textsc{RiemannOpt}. High values of $|g'(\alpha)|$ indicate regions of the path where the gradients of the model are rapidly changing, i.e. regions where the image becomes perceptible to the model.}
    \label{fig:optimal_points}
\end{figure}

\section{Results and Discussion} 
In this section, we compare the sampling points chosen by \textsc{RiemannOpt} to the linear schedules, followed by qualitative and quantitative evaluation against the baselines. In the case of BlurIG, the sample points chosen by \textsc{RiemannOpt} highly differ from the linearly spaced samples, as depicted in Figure \ref{fig:optimal_points}. Every path starts with an information-less baseline image, $x'$, and gradually gains perceptible features as it moves towards the input image, $x$. Along the path, when the image becomes perceptible, the gradients rapidly change, resulting in large values of $|g'(\alpha)|$. For BlurIG, the image features become perceptible at the end of the path when most of the sharpening occurs, $\alpha \approx 0.8$. For IG and GIG, the image becomes perceptible as soon as its brightness crosses a certain threshold, $\alpha \approx 0.1$. The abnormal spike at the end of GIG's curve is due to its adaptive path mechanism.

\begin{figure*}[!h]
\centering
\hspace*{-0.75cm}
\begin{tikzpicture}
\pgfplotsset{
    IG/.style={color=red!90!black,mark=triangle,dotted,mark options={solid,fill=red!90!black}},
    IGOurs/.style={color=red!90!black,mark=square*},
    BlurIG/.style={color=blue!70!black,mark=triangle,dotted,mark options={solid,fill=blue!70!black}},
    BlurIGOurs/.style={color=blue!70!black,mark=square*},
    GIG/.style={color=green!60!black,mark=triangle,dotted,mark options={solid,fill=green!60!black}},
    GIGOurs/.style={color=green!60!black,mark=square*},
}

\centering
\begin{groupplot}[
    group style={
        group size=2 by 1,
        horizontal sep=0.1\textwidth,
        vertical sep=0.0556\textheight,
    },
    width=0.56\textwidth,
    height=0.24\textheight,
    xlabel={Samples},
    xlabel style={yshift=+2mm},
    xmin=15, xmax=130,
    xtick={16,32,64,128},
    xticklabels={$2^4$,$2^5$,$2^6$,$2^7$},
    xmajorgrids=true,
    ymajorgrids=true,
    xminorgrids=true,
    yminorgrids=true,
    xmode=log,
    log basis x=2,
    tick label style={font=\tiny},
    label style={font=\footnotesize},
    title style={font=\small, at={(0.5,0.89)}, yshift=2mm},
    every axis plot/.append style={line width=0.8pt},
    legend cell align=left,
    clip=false,
    legend style={
        at={(0.5,1.05)},
        xshift=10.67cm,
        yshift=1.15cm,
        legend columns=6,
        /tikz/every even column/.append style={column sep=0.35cm}
    },
]

\nextgroupplot[
    ymin=0.25, ymax=0.45,
    title={Insertion Score ($\uparrow$)},
]
\addplot[IG] coordinates {
    (16,0.36155823) (32,0.36369896) (64,0.36567894) (128,0.368)
};
\addlegendentry{IG}
\addplot[IGOurs] coordinates {
    (16,0.4396924) (32,0.44105795) (64,0.44110486) (128,0.445)
};
\addlegendentry{IG + Ours}
\addplot[BlurIG] coordinates {
    (16,0.3026216) (32,0.32038304) (64,0.32449973) (128,0.32510218)
};
\addlegendentry{BlurIG}
\addplot[BlurIGOurs] coordinates {
    (16,0.32414642) (32,0.32445028) (64,0.32535937) (128,0.3402909)
};
\addlegendentry{BlurIG + Ours}
\addplot[GIG] coordinates {
    (16,0.28022137) (32,0.2899874) (64,0.29048866) (128,0.2871344)
};
\addlegendentry{GIG}
\addplot[GIGOurs] coordinates {
    (16,0.28214455) (32,0.2951737) (64,0.29467) (128,0.2920428)
};
\addlegendentry{GIG + Ours}

\nextgroupplot[
    ymin=0.3, ymax=0.55,
    title={Normalized Insertion Score ($\uparrow$)},
]
\addplot[IG] coordinates {
    (16,0.42238092) (32,0.4247464) (64,0.42705083) (128,0.429)
};
\addplot[IGOurs] coordinates {
    (16,0.5161938) (32,0.5177384) (64,0.51769274) (128,0.51782)
};
\addplot[BlurIG] coordinates {
    (16,0.35304818) (32,0.37498865) (64,0.3800835) (128,0.38075045)
};
\addplot[BlurIGOurs] coordinates {
    (16,0.37986144) (32,0.38029265) (64,0.3813283) (128,0.39960468)
};
\addplot[GIG] coordinates {
    (16,0.32821578) (32,0.33898976) (64,0.33903813) (128,0.33529568)
};
\addplot[GIGOurs] coordinates {
    (16,0.33044213) (32,0.34356112) (64,0.34334726) (128,0.34008495)
};

\end{groupplot}

\end{tikzpicture}
\vspace{-0.7cm}
\caption{We compare \textsc{RiemannOpt} against the baseline methods using the Insertion Score and Normalized Insertion Score. We observe noticeable improvement for BlurIG and IG.}
\label{fig:metrics-comparison}
\end{figure*}
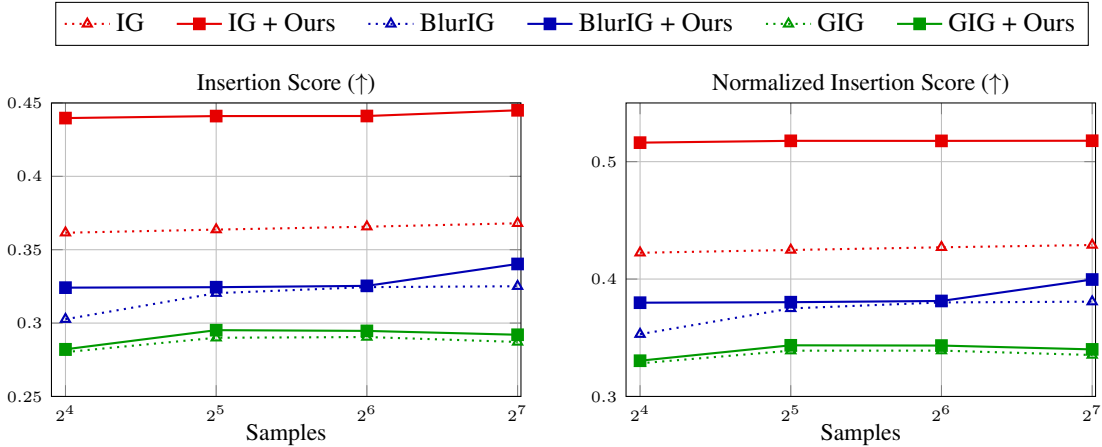

\textsc{RiemannOpt} always reduces the relative error and improves metric scores across all methods and sample counts as depicted in Table \ref{tab:relative-error} and Figure \ref{fig:metrics-comparison} respectively, with a noticeable enhancement for BlurIG. On the other hand, the improvement in GIG is not very significant. The path of GIG is theoretically fixed for chosen model. However, due to the employment of an adaptive path, its practical implementation is highly dependent on the number of samples as well as the location of the samples, unlike BlurIG and IG. In the derivation \ref{derivation} of \textsc{RiemannOpt}, we assumed that the path function was constant and independent of the sample points. The practical implementation of GIG breaks this assumption; this is a possible explanation for why GIG is not as improved by \textsc{RiemannOpt} as the other methods are. In terms of relative error, \textsc{RiemannOpt} significantly reduces the number of samples while maintaining comparable performance. Specifically, BlurIG + \textsc{RiemannOpt} achieves similar results with $16$ samples as BlurIG with $64$ samples. Additionally, BlurIG + \textsc{RiemannOpt} with $16$ samples performs comparably to BlurIG with $32$ samples, and GIG + \textsc{RiemannOpt} matches the performance of GIG with $128$ samples using just $16$ samples. This makes \textsc{RiemannOpt} highly functional for computationally constrained environments.

\begin{table}[!h]
\caption{Relative Error ($\downarrow$) across different methods}
\label{tab:relative-error}
\centering
\renewcommand{\arraystretch}{1.5} 
\setlength{\tabcolsep}{10pt} 
\begin{tabular}{lcccc}
\hline\hline 
Method & 16 Samples & 32 Samples & 64 Samples & 128 Samples \\
\hline\hline 
IG & 0.708 & 0.374 & 0.166 & 0.066 \\
\textbf{IG + \textsc{RiemannOpt}} & \textbf{0.404} & \textbf{0.223} & \textbf{0.123} & \textbf{0.065} \\
\hline
BlurIG & 0.886 & 0.554 & 0.268 & 0.114 \\
\textbf{BlurIG + \textsc{RiemannOpt}} & \textbf{0.269} & \textbf{0.123} & \textbf{0.058} & \textbf{0.041} \\
\hline
GIG & 0.786 & 0.788 & 0.725 & 0.612 \\
\textbf{GIG + \textsc{RiemannOpt}} & \textbf{0.666} & \textbf{0.731} & \textbf{0.711} & \textbf{0.610} \\
\hline
\end{tabular}
\end{table}

\begin{figure}[!h]
    \centering
    \includegraphics[width=1\linewidth]{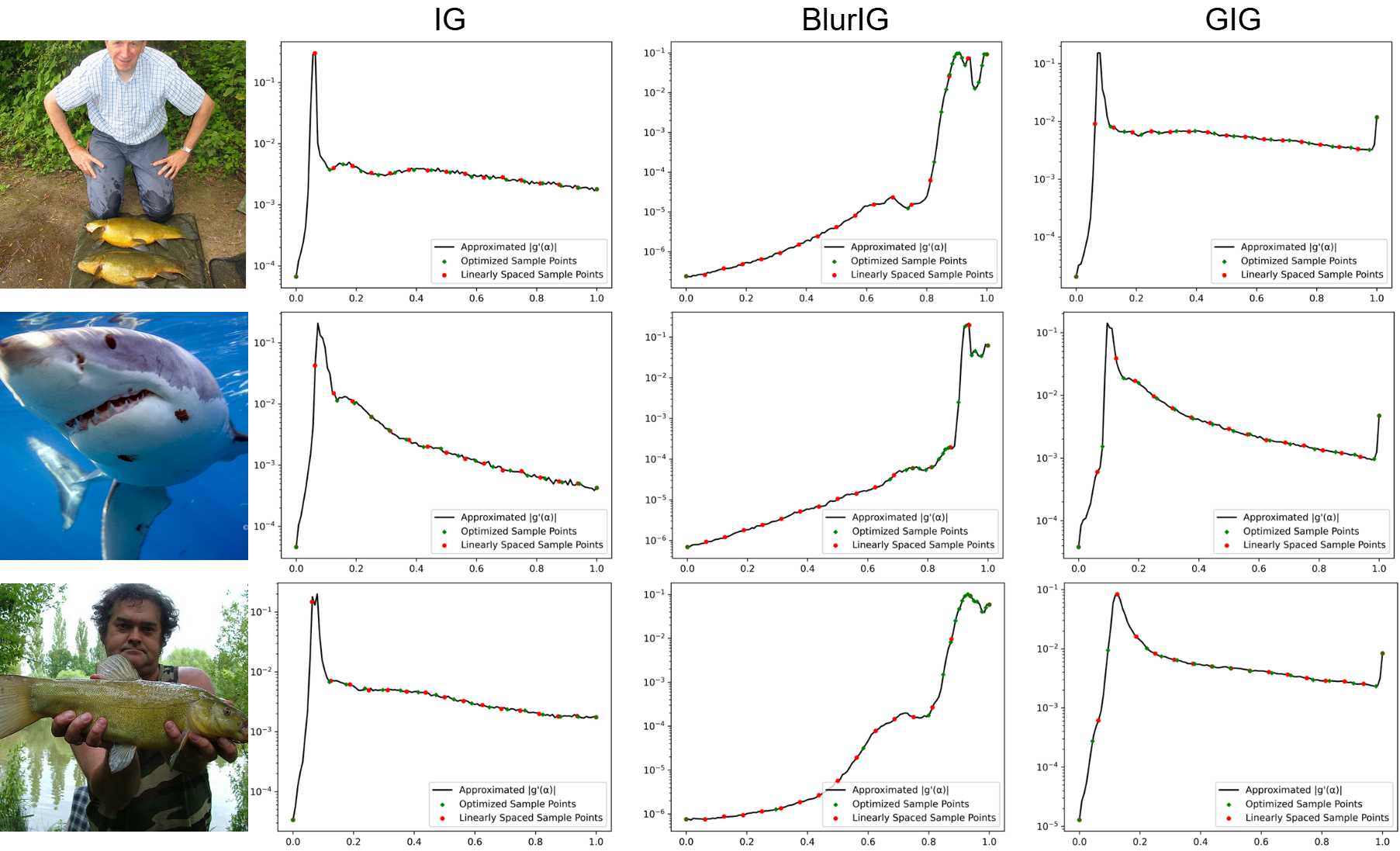}
    \caption{$|g'(\alpha)|$ plots for individual images. The shapes of the plots and the generated optimal points are similar, highlighting the generalizability of the algorithm to images across the dataset.}
    \label{fig:examples}
\end{figure}

The reader may question to what extent the optimal points, approximated for the entire dataset, generalize for each individual image. This can be verified by applying \textsc{RiemannOpt} for several individual images and qualitatively inspecting how much the resultant points vary for different images. As shown in Figure \ref{fig:examples}, for ImageNet we observed that the set of points generated was roughly the same. One would expect the same trend even with different datasets since the characteristics of $|g'(\alpha)|$ are primarily determined by the path construction and model's training procedure rather than the image itself. InceptionV3 was trained with augmentation techniques including random brightness shifts which increased its robustness to dark images, making the image perceptible to the model at around $\alpha \approx 0.1$, as previously mentioned. Future work could investigate how changing the in-training augmentation parameters would affect the shape of the graph and consequently \textsc{RiemannOpt's} chosen points, including blurring augmentations. One would expect the rise in $|g'(\alpha)|$ for BlurIG's path to occur earlier since the model would be better at identifying blurry images, similar to the effect caused by brightness augmentation.

%
\section{Conclusion}
In this paper, we present \textsc{RiemannOpt}, a highly efficient framework designed to optimize sample points in Riemann Sums for the computation of Integrated Gradients. Both qualitative and quantitative results demonstrate that \textsc{RiemannOpt} effectively minimizes numerical errors in saliency maps and achieves improved Insertion Scores by up to $20\%$, thereby enhancing the accuracy and reliability of attribution maps. \textsc{RiemannOpt} is adaptable, extending its applicability to any multi-dimensional line integral computation, including derivatives of Integrated Gradients such as BlurIG and GIG. Additionally, it enables users to curtail computational costs by up to fourfold, significantly boosting efficiency. Opportunities for future work include extending \textsc{RiemannOpt} to further improve its suitability for Integrated Gradient methods that employ adaptive paths.

\begin{ack}
We would like to thank Gaurav Kumar Nayak, Aayan Yadav, Shweta Singh, Anupriya Kumari and Devansh Bhardwaj for their insights on the paper writing. We would also like to thank all members of the Data Science Group of IIT Roorkee for their invaluable support.
\end{ack}

\newpage
\bibliographystyle{plainnat}
\bibliography{references}








\end{document}